\documentclass[runningheads]{llncs}
\usepackage{times}  
\usepackage{helvet}  
\usepackage{courier}  
\usepackage{url}  
\usepackage{graphicx}  
\usepackage{amsmath}
\usepackage{amsfonts}
\usepackage[utf8]{inputenc}
\begin{document}
\title{Basic and Depression Specific Emotion Identification in Tweets: Multi-label Classification Experiments}
%
%
\author{Nawshad Farruque \and
Chenyang Huang \and
Osmar Zaiane \and
Randy Goebel}
\authorrunning{Farruque, N. et al.}
%
\institute{Dept. of Computing Science, University of Alberta, Canada \\
\email{\{nawshad,chuang8,zaiane,rgoebel\}@ualberta.ca}}

%
\maketitle              
\begin{abstract}
In this paper, we present empirical analysis on basic and depression specific multi-emotion mining in Tweets with the help of state of the art multi-label classifiers. We choose our basic emotions from a hybrid emotion model consisting of the common emotions from four highly regarded psychological models of emotions. Moreover, we augment that emotion model with new emotion categories because of their importance in the analysis of depression. Most of those additional emotions have not been used in previous emotion mining research.  Our experimental analyses show that a cost sensitive RankSVM algorithm and a Deep Learning model are both robust, measured by both Macro F-measures and Micro F-measures. This suggests that these algorithms are superior in addressing the widely known data imbalance problem in multi-label learning. Moreover, our application of Deep Learning performs the best, giving it an edge in modeling deep semantic features of our extended emotional categories.
\end{abstract}

\section{Introduction}
Mining multiple human emotions can be an interesting area of research since human emotions tend to co-occur \cite{mill2018role}. For example, most often human emotions such as $joy$ and $surprise$ tend to occur together than just $joy$ or just $surprise.$ (See Table \ref{tab:multi-label-tweets} for some examples from our dataset).
Also, identifying these co-occurrences of emotions and their compositionality can provide insight for fine grained analysis of emotions in various mental health problems. There are very few literature that have explored multi-label emotion mining from text \cite{bhowmick2009reader,luyckx2012fine,shahrakilexical}. With the increasing use of social media, where people share their day to day thoughts and ideas, it is easier to capture the presence of different emotions in their 
posts. 

Thus the main focus of this research is to provide insights on identifying multi-emotions from social media posts such as Tweets. To compile a list of emotions we want to identify, we have used a mixed emotion model \cite{shahrakilexical} which is based on four distinct and widely used emotion models used in psychology. Furthermore, we augment this emotion model with more emotions that are deemed useful for a depression identification task we intend to pursue later. Here we separate our experiments into two: one for a smaller emotion model (using nine basic human emotions), and another for an augmented emotion model (using both basic and depression related human emotions). 

We present a detailed analysis of the performance of several state of the art algorithms used in multi-label text mining tasks on these two sets of data, both of which have varying degree of data imbalance. 

\begin{table} [!ht]
\begin{center}
\begin{tabular}{|p{5cm}|l|}
\hline
\textbf{Tweets} & \textbf{Labels} \\
\hline
``Feel very annoyed by everything and I hope to leave soon because I can't stand this anymore'' & angry, sad \\
\hline
``God has blessed me so much in the last few weeks I can't help but smile''  &joy, love \\
\hline
\end{tabular}
\caption{\label{tab:multi-label-tweets}Example Tweets with multi-label emotions}
\end{center}
\end{table}

\subsection{Emotion Modeling} 
In affective computing research, the most widely accepted model for emotion is the one suggested by \cite{ekman1992argument}. That model consists of six emotions: anger, disgust, fear, joy, sadness and surprise. Here we adopt an emotional model more recently proposed in \cite{shahrakilexical}, which is a generalization from these models, augmented with a small number of additional emotions: love, thankfulness and guilt, all of which are relevant to our study. We further seek to confirm emotions such as betrayal,  frustration, hopelessness, loneliness, rejection, schadenfreude and self loathing; any of these could contribute to the identification of a depressive disorder 
\cite{abramson1989hopelessness,pietraszkiewicz2015link}. Our mining of these emotions, with the help of RankSVM and an attention based deep learning model, is a new contribution that  has been previously lacking 
\cite{mohammad2012emotional,hasan2014using}.

\subsection{Multi-label Emotion Mining Aproaches}
Earlier research in this area has employed learning algorithms from two broad categories: (1) Problem Transformation and (2) Algorithmic Adaptation. A brief description of each of these follows in the next subsections. 

\subsection{Problem Transformation Methods} 
In the problem transformation approach (PTM), multi-label data is transformed into single label data, and a series of single label (or binary) classifiers are trained for each label. Together, they predict multiple labels (cf. details provided in Section \ref{sect:baseline}). This method is often called a ``one-vs-all'' model. The problem is that these methods do not consider any correlation amongst the labels. This problem was addressed by a model proposed by \cite{bhowmick2009reader}, which uses label powersets (LP) to learn an ensemble of k-labelset classifiers (RAKEL) \cite{tsoumakas2011random}. Although this classifier method respects the correlation among labels, it is not robust against the data imbalance problem, which is an inherent problem in multi-label classification, simply because of the typical uneven class label distribution.

\subsection{Algorithmic Adaptation Methods} 
The alternative category to PTMs are the so-called algorithmic adaptation methods (AAMs), where a single label classifier is modified to do multi-label classification. Currently popular AAMs are based on trees, such as the classic C4.5 algorithm adapted for multi-label task \cite{clare2001knowledge}, probabilistic models such as \cite{ghamrawi2005collective}, and neural network based methods such as, BP-MLL \cite{zhang2006multilabel}. However, as with PTMs, these AAM methods are also not tailored for imbalanced data learning, and fail to achieve good accuracy with huge multi-label dataset where imbalance is a common problem.

In our approach, we explore two state of the art methods for multi-label classification. One is a cost sensitive RankSVM and the other is a deep learning model based on Long Short Term Memory (LSTM) and Attention. The former is an amalgamation of PTM and AAMs, with the added advantage of large margin classifiers.  This provides an edge on learning based on huge imbalanced multi-label data, while still considering the label correlations. The later is a purely AAM approach, which is able to more accurately capture latent semantic structure of Tweets. In Sections \ref{sect:baseline} and \ref{sect:expmodels} we provide the technical details of our baseline and experimental models. 

\section{Baseline Models}
\label{sect:baseline}
According to the results in \cite{shahrakilexical}, good accuracy can be achieved with a series of Na\"{\i}ve Bayes (NB) classifiers (also known as a one-vs-all classifier), where each of them are trained on balanced positive and negative samples (i.e., Tweets, represented by bag-of-words features) for each class; especially with respect to  other binary classifiers (e.g., Support Vector Machines (SVMs)). To recreate this baseline, we have implemented our own ``one-vs-all'' method. To do so, we transform our multi-label data into sets of single label data, then train separate binary NB classifiers for each of the labels. An NB classifier $NB_i$ in this model uses emotion $E_i$ as positive sample and all other emotion samples as negative samples, where $i$ is a representative index of an emotion. We next concatenate the binary outputs of these individual classifiers to get the final multi-label output. It is also to be noted that previous closely related research \cite{luyckx2012fine,bhowmick2009reader} used simple SVM and RAKEL, which are not robust against data imbalance and did not look at short text, such as Tweets for multi-label emotion mining. On the other hand, \cite{shahrakilexical} had a focus on emotion mining from Tweets but their methods were multi-class, unlike us, where we are interested in multi-label emotion mining.


\section{Experiment Models}
\label{sect:expmodels}

\subsection{A Cost Sensitive RankSVM Model}
\label{ssect:ranksvm}
Conventional multi-label classifiers learn a mapping function, $h:X \rightarrow 2^q$ from a $D$ dimensional feature space, 
$X \in \mathbb{R}^D$ to the label space $Y \subseteq {\{0,1\}}^{q}$, where $q$ is the number of labels. 
A simple label powerset algorithm (LP) considers each distinct combination of labels (also called labelsets) that exist in the training data as a single label, thus retaining the correlation of labels. 
In multi-label learning, some labelsets occur more frequently than others, and traditional SVM algorithms perform poorly in these scenarios. 
In many information retrieval tasks, the RankSVM algorithm is widely used to learn rankings for documents, given a query. This idea can be generalized to multi-label classification, where the relative ranks of labels is of interest. \cite{cao2016cost} have proposed two optimized versions of the RankSVM algorithm \cite{joachims2002optimizing}: one is called RankSVM(LP), which not only incorporates the LP algorithm but also associates misclassification cost $\lambda_i$ with each training instance. This misclassification cost is higher for the label powersets that have smaller numbers of instances, and is automatically calculated based on the distribution of label powersets in the training data. To further reduce the number of generated label powersets (and to speed up processing), they proposed another version of their algorithm called RankSVM (PPT), where labelsets are pruned by an a priori threshold based on properties of the data set.

\subsection{A Deep Learning Model}
\label{ssect:deeplearn}
Results by \cite{zhou2015c} showed that a combination of Long Short Term Memory (LSTM) and an alternative to LSTM, Gated Recurrent Unit layers (GRU), can be very useful in learning phrase level features which achieved very good accuracy in text classification.  \cite{lin2017structured} achieved state of the art in sentence classification with the help of bidirectional LSTM (bi-LSTM) with a self-attention mechanism. 
Here we adopt \cite{lin2017structured}'s model 
and further enable this model for multi-label classification by using a suitable loss function and a thresholded softmax layer to generate multi-label output. We call this model LSTM-Attention (LSTM-att), as shown in Figure~\ref{fig:Attention-model};  $w_i$ is the word embedding (which can be either one-hot bag-of-words or dense word vectors), $h_i$ is the hidden states of LSTM at time step $i$, and the output of this layer is fed to the Self Attention (SA) layer. SA layer's output is then sent to a linear layer, which translates the final output to a probability of different labels (in this case emotions) with the help of softmax activation. Finally, a threshold is applied on the softmax output to get the final multi-label predictions.

\subsection{Loss Function}
\begin{figure}[!ht]
  \centering
  \includegraphics[width=0.78\textwidth]{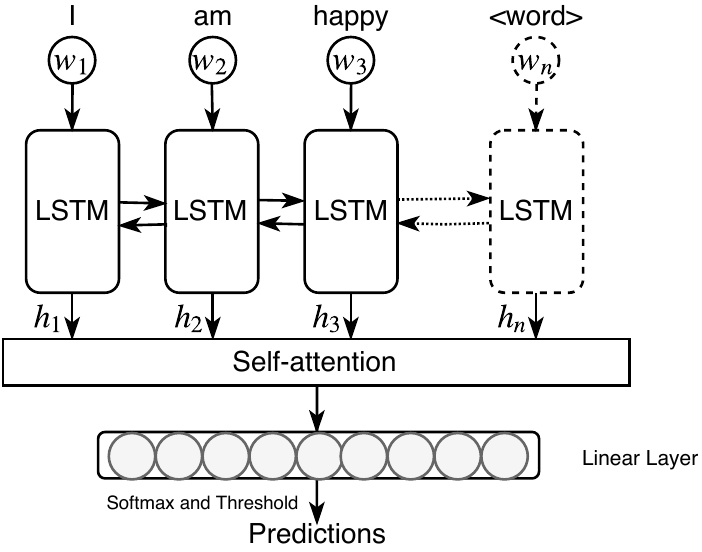}
  \caption{\label{fig:Attention-model}LSTM-Attention model for multi-label classification}
\end{figure}

The choice of a loss function is important in this context, and for multi-label classification task \cite{liu2017deep}, it has shown that Binary Cross Entropy (BCE) Loss over sigmoid activation is very useful. Our use of a BCE objective can be formulated as follows:

\begin{equation}
\begin{split}
{\text{minimize}} ~  \frac{1}{n}\sum_{i=1}^{n}\sum_{l=1}^{L} [y_{il}log(\sigma(\hat{y_{il}})) \\
+ (1 - y_{il})log(1 - \sigma(\hat{y_{il}}))]
\end{split}
\end{equation}
\noindent where $L$ is the number of labels, $n$ is the number of samples, $y_i$ is the target label, $\hat{y_i}$ is the predicted label from last linear layer (see in Figure \ref{fig:Attention-model}) and $\sigma$ is a sigmoid function, $\sigma(x) = \frac{1}{1 + e^{-x}}$.

\subsection*{Thresholding}
The output of the last linear layer is a k-dimensional vector, where $k$ is the number of different labels of the classification problem. For our task we use the \textit{softmax} function to normalize each $\hat{y_i}$ within the range of $(0,1)$ as following,

\begin{equation}
\hat{y_i}' = {\frac {e^{\hat{y_i}}} {\sum _{i=1}^{k}e^{\hat{y_i}}}}    
\end{equation}
Since we are interested in more than one label, we use a threshold and consider those labels with predicted probability beyond that threshold. Let the threshold be $t$, hence the final prediction for each label $p_i$ is 
\begin{equation}
p_i = 
\begin{cases}
    1, &  {y_i}' > t\\ 
    0, & else
\end{cases}
\end{equation}

\noindent 
To adjust the thresholds for the LSTM-Att model, we use a portion of our training data as an evaluation set. Based on the evaluation set we find the threshold, $t = 0.3$ provides the best results.

\section{Experiments}
\label{sect:exp}
We use the well-known bag-of-words (BOW) 
and pre-trained word embedding vectors (WE) 
as our feature sets, for two RankSVM algorithms: RankSVM-LP and RankSVM-PPT, a one-vs-all Naive Bayes (NB) classifier and a deep learning model (LSTM-Att). We name our experiments with algorithm names suffixed by feature names, for example RankSVM-LP-BOW names the experiment with the RankSVM Label Powerset function on a bag-of-words feature set. We run these experiments on our two sets of data, and we use the RankSVM implementation provided by \cite{cao2016cost}. For multi-label classification, we implement our own one-vs-all model using a Python library named sci-kit learn and its implementation of multinomial NB. We implement our deep learning model in pyTorch\footnote{http://pytorch.org/}, a deep learning framework released by Facebook. In the next section, we present a detailed description of the datasets, data collection, data pre-processing, feature sets extraction and evaluation metrics.

\subsection{Data Set Preparation}
\label{ssect:datacol}
We use two sets of multi-label data \footnote{We intend to release our dataset online upon publication of this paper.}. Set 1 (we call it 9 emotion data) consists of Tweets only from the ``Clean and Balanced Emotion Tweets'' (CBET) dataset provided by \cite{shahrakilexical}. It contains 3000 Tweets from each of nine emotions, plus 4,303 double labeled Tweets (i.e., Tweets which have two emotion labels), for a total of 31,303 Tweets. To create Set 2 (we call it 16 emotion data), we add extended emotions Tweets (having single and double labels) with Set 1 data adding up to total 50,000 Tweets. We used Tweets having only one and two labels because this is the natural distribution of labels in our collected data. Since previous research showed that hashtag labeled Tweets are consistent with the labels given by human judges \cite{mohammad2015using}, these Tweets were collected based on relevant hashtags and key-phrases \footnote{We use specific key-phrases to gather Tweets for specific emotions, e.g. for loneliness, we use, ``I am alone'' to gather more data for our 
extra emotion data collection process.}.
Table \ref{tab:extra-emo} lists the additional emotions we are interested in. Our data collection process is identical to \cite{shahrakilexical}, except that we use the Twitter API and key-phrases along with hashtags. In this case, we gather Tweets with these extra emotions between June, 2017 to October, 2017. The statistics of gathered Tweets are presented in Table \ref{tab:extra-emo-counts}. Both of our data sets have the following characteristics after preprocessing:

\begin{itemize}
\item All Tweets are in English.
\item All the letters in Tweets are converted to lowercase. 
\item White space characters, punctuations and stop words are removed. 
\item Duplicate Tweets are removed.
\item Incomplete Tweets are removed.
\item Tweets shorter than 3 words are removed.
\item Tweets having more than 50\% of its content as name mentions are removed. 
\item URLs are replaced by 'url'.
\item All name mentions are replaced with `@user'.
\item Multi-word hashtags are decomposed in their constituent words.
\item Hashtags and key-phrases corresponding to emotion labels are removed to avoid data overfitting.
\end{itemize}

\noindent Finally, we use 5 fold cross validation (CV) (train 80\% - test 20\% split). In each fold we further create a small validation set (10\% of the training set) and use it for parameter tuning in Rank SVM and threshold finding for LSTM-Att. Our baseline model does not have any parameters to tune. Finally, our results are averaged on the test set in the 5 Fold CV based on the best parameter combination that was found in each of the folds validation set and we report that. Using this approach, we find that threshold = 0.3 is generally better. We do not do heavy parameter tuning in LSTM-att. Also, we use the Adam Optimizer with 0.001 learning rate.


\begin{table}[!ht]
\begin{center}
\begin{tabular}{|l|p{50mm}|}
\hline \bf Emotion & \bf List of Hashtags \\ \hline
betrayed & \#betrayed  \\
frustrated   & \#frustrated, \#frustration  \\
hopeless & \#hopeless, \#hopelessness, no hope, end of everything  \\
loneliness & \#lonely, \#loner, i am alone \\
rejected & \#rejected, \#rejection, nobody wants me, everyone rejects me \\
schadenfreude & \#schadenfreude \\
self loath & \#selfhate, \#ihatemyself, \#ifuckmyself, i hate myself, i fuck myself\\
\hline
\end{tabular}
\end{center}
\caption{\label{tab:extra-emo} New emotion labels and corresponding hashtags and key phrases }
\end{table}

\begin{table}[!ht]
\begin{center}
\begin{tabular}{|l|c|}
\hline \bf Emotion & \bf Number of Tweets \\ \hline
betrayed &   1,724\\
frustrated   &  4,424 \\
hopeless &   3,105\\
loneliness &  4,545\\
rejected &  3,131\\
schadenfreude &  2,236\\
self loath & 4,181\\
\hline \bf Total & \bf 23,346 \\ \hline
\end{tabular}
\end{center}
\caption{\label{tab:extra-emo-counts} Sample size for each new emotions (after cleaning)}
\end{table}

\subsection{Feature Sets}
We create a vocabulary of 5000 most frequent words which occur in at least three training samples. If we imagine this vocabulary as a vector where each of its indices represent each unique word in that vocabulary, then a bag-of- words (BOW) feature can be represented by marking as 1 those indices whose corresponding word matches with a Tweet word. To create word embedding features we use a 200 dimensional GloVe \cite{pennington2014glove} word embeddings trained on a corpus of 2B Tweets with 27B tokens and 1.2M vocabulary\footnote{https://nlp.stanford.edu/projects/glove/}. We represent each Tweet with the average word embedding of its constituent words that are also present in the pre-trained word embeddings. We further normalize the word embedding features using min-max normalization. 

\subsection{Evaluation Metrics}
We report some metrics for our data imbalance measurements 
to highlight the effect of the analysed approaches in multi-label learning.
\subsection{Quantifying Imbalance in Labelsets}
\begin{figure}[!ht]
\centering
\includegraphics[width=0.49\textwidth]{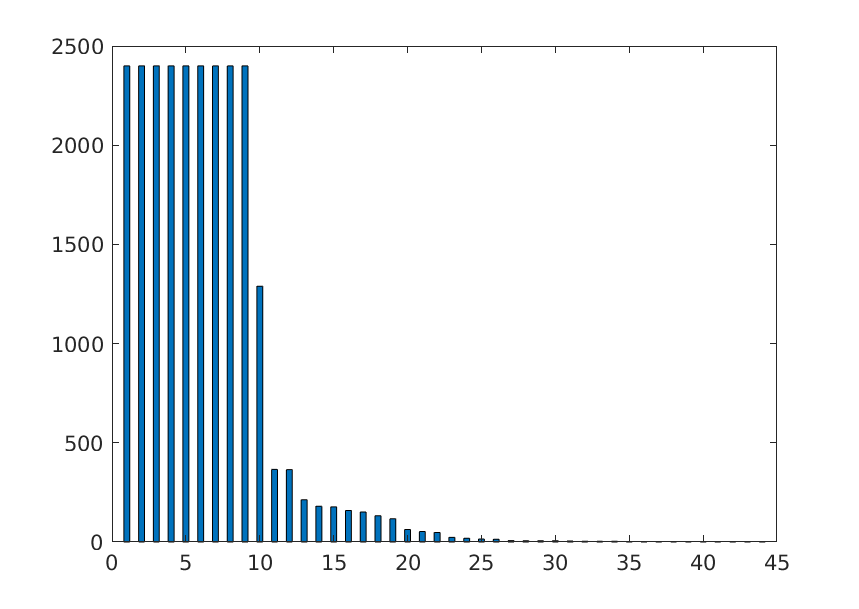}
\includegraphics[width=0.49\textwidth]{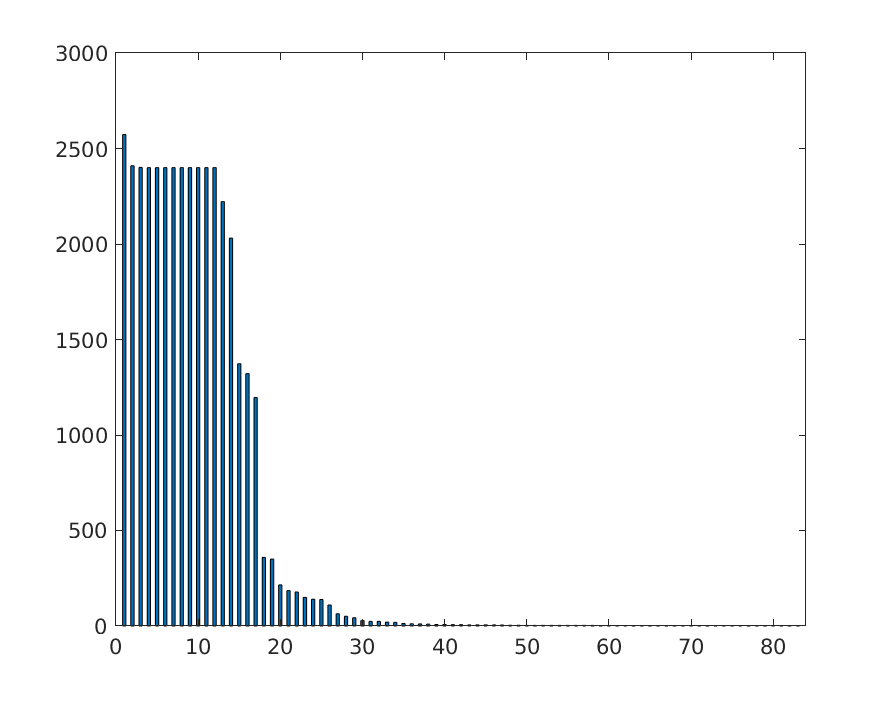}
\caption{\label{fig:labelsetimbal9} Labelset imbalance: histogram for 9 emotion dataset on the left and histogram for 16 emotion dataset on the right}
\end{figure}

We use the same idea as mentioned in \cite{cao2016cost} for data imbalance calculation in labelsets, inspired by the idea of kurtosis. The following equation is used to calculate the imbalance in labelsets, 

\begin{equation}
ImbalLabelSet = \frac{\sum_{i=1}^{l}(L_i-L_{max})^4}{(l -1)s^4}
\end{equation}

where,
\begin{equation}
s = \sqrt{\frac{1}{l} \sum_{i=1}^l(L_i-L_{max})}
\end{equation}

\noindent here, $l$ is the number of all labelsets, $L_{i}$ is the number of samples in the $i-th$ labelset, $L_{max}$ is the number of samples with the labelset having the maximum samples. The value of $LabelSetImbal$ is a measure of the distribution shape of the histograms depicting labelset distribution. A higher value of $LabelSetImbal$ indicates a larger imbalance, and determines the ``peakedness'' level of the histogram (see Figure \ref{fig:labelsetimbal9} 
), where $x$ axis depicts the labelsets and $y$ axis denotes the counts for them. The numeric imbalance level in labelsets is presented in Table \ref{tab:imbaldeg}; we can see that the 16 emotion dataset is more imbalanced than the 9 emotion dataset. 

\begin{table}[!ht]
\begin{center}
\begin{tabular}{|l|c|c|}
\hline \bf Dataset & \bf Labelset Imbal. & \bf Label Imbal. \\ \hline
9 Emotion Data & 27.95 & 37.65 \\
16 Emotion Data & 52.16 & 69.93 \\
\hline
\end{tabular}
\end{center}
\caption{\label{tab:imbaldeg} Degree of Imbalance in Labelset and Labels }
\end{table}

\subsection{Mic/Macro F-Measures}
We use label based Mic/Macro F-Measures to report the performance of our classifiers in classifying each labels. See Equations \ref{eq:micro-fm} and \ref{eq:macro-fm}.

\begin{equation}
\label{eq:micro-fm}
\begin{split}
    Micro-FM = F-Measure(\sum_{j=1}^{l}TP_{l}, \sum_{j=1}^{l}FP_{l},\sum_{j=1}^{l}TN_{l},\sum_{j=1}^{l}FN_{l})
\end{split}
\end{equation}

\begin{equation}
\label{eq:macro-fm}
\begin{split}
    Macro-FM = \frac{1}{l}\sum_{j=1}^{l}F-Measure(TP_{j},FP_{j} ,TN_{j},FN_{j})
\end{split}
\end{equation}

where, $l$ is is the number of labels, and $F-measure$ is a standard F1-score and $TP, FP, TN, FN$ are, True Positive, False Positive, True Negative and False Negative labels respectively. 

Micro-FM provides the performance of our classifiers by taking into account the imbalance in the dataset unlike Macro-FM. 

\section{Results Analysis}
\begin{figure}[!ht]
\centering
\includegraphics[width=0.60\textwidth]{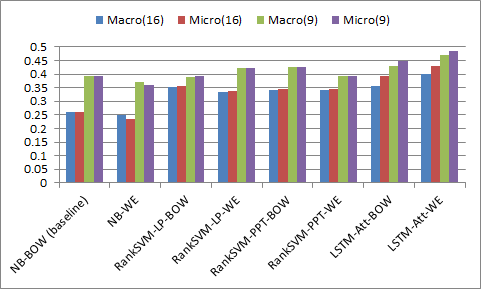}
\caption{\label{fig:histogram} Comparative results on 9 and 16 emotion datasets}
\end{figure}

Overall, LSTM-Attention with word embedding features (LSTM-Att-WE) performs the best in terms of Macro-FM and Micro-FM measures, compared with the baseline multi-label NB and best performing RankSVM models averaged across 9 emotion and 16 emotion datasets. These results are calculated according to Equation \ref{eq:6}, 
where, $BestFM_{i}$ refers to the model with best F-measure value (either Macro or Micro),  $i$ indicates the dataset and can be either 9 or 16, 
$AVG$ denotes the function that calculates average, the sub-script index refers to our $9$ or $16$ emotion dataset, and other variables are self explanatory. To compare with RankSVM, we use best RankSVM F-measures instead of $BaselineFM$ in the above equation 
In Macro-FM measure LSTM-Att-WE shows a $37\%$ increase with regard to baseline NB-BOW and $18\%$ increase with regard to best RankSVM models. But in the Micro-FM measure, LSTM-Att achieves $44\%$ increase with regard to baseline NB and $23\%$ increase with regard to best RankSVM models.  The percentage values were rounded to the nearest integers. It is worth noting that a random assignment to classes would result in an accuracy of $11\%$ in the case of the $9$ emotions $(1/9)$ and $6\%$ in the case of the $16$ emotion dataset $(1/16)$. The following sections show analyses based on F-measures, Data Imbalance and Confusion Matrices (see Tables \ref{tab:9emotionResults}, \ref{tab:16emotionResults} and Figure \ref{fig:histogram}) .

\begin{equation}
\begin{split}
\label{eq:6}
FM-Inc = AVG(\frac{BestFM_{16} -  BaselineFM_{16}}{BaselineFM_{16}}, \\ \frac{BestFM_{9} - BaselineFM_{9}}{BaselineFM_{9}}) \times 100
\end{split}
\end{equation}

\begin{table}[!ht]
\begin{center}
\begin{tabular}{|l|l|l|}
\hline
Models & Macro-FM & Micro-FM \\
\hline
NB-BOW (baseline) 	&0.3915		&0.3920	\\	
NB-WE 	&0.3715		&0.3617	\\	
RankSVM-LP-BOW	&0.3882			&0.3940	\\
RankSVM-LP-WE		&0.4234			&0.4236	\\
RankSVM-PPT-BOW	&0.4275			&0.4249	\\
RankSVM-PPT-WE	&0.3930			&0.3920	\\
LSTM-Att-BOW		&0.4297			&0.4492	\\
LSTM-Att-WE		&\textbf{0.4685}			&\textbf{0.4832}	\\
\hline
\end{tabular}
\caption{\label{tab:9emotionResults}Results on 9 Emotion Dataset}
\end{center}
\end{table}


\begin{table}[!ht]
\begin{center}
\begin{tabular}{|l|l|l| }
\hline
Models & Macro-FM & Micro-FM \\
\hline
NB-BOW (baseline) 	&0.2602		&0.2608	\\	
NB-WE 	&0.2512		&0.2356	\\	
RankSVM-LP-BOW	&0.3523			&0.3568	\\
RankSVM-LP-WE		&0.3342			&0.3391	\\
RankSVM-PPT-BOW	&0.3406			&0.3449	\\
RankSVM-PPT-WE	&0.3432			&0.3469	\\
LSTM-Att-BOW		&0.3577			&0.3945	\\
LSTM-Att-WE		&\textbf{0.4020}			&\textbf{0.4314}	\\
\hline
\end{tabular}
\caption{\label{tab:16emotionResults}Results on 16 Emotion Dataset}
\end{center}
\end{table}

\subsection{Performance with regard to F-measures}
Overall, in all models, the Macro-FM and Micro-FM values are very close to each other (see Table \ref{tab:9emotionResults} and \ref{tab:16emotionResults}) indicating that all of the models have similar performance in terms of both most populated and least populated classes.

\subsection{Performance with regard to Data Imbalance}
Performance of all the models generally drops as the imbalance increases. For this performance measure, we take average F-measures (Macro or Micro) across WE and BOW features for NB and DL; for RankSVM the values are averaged over two types of RankSVM models as well (i.e., LP and PPT). 
See Equation \ref{eq:7}, 
where, the function, performance drop, $PD(model)$, takes any model and calculates the decrease or drop of F-measures (either Macro or Micro) averaged over WE and BOW features. For RankSVM, we input RankSVM-PPT and RankSVM-BOW, and take the average to determine the performance drop in overall RankSVM algorithm based on Equation,
\ref{eq:8}.
We observe that LSTM-Att's performance drop for Macro-FM is $15\%$, and Micro-FM is $11\%$ with respect to $9$ emotions (less imbalanced) to $16$ emotion data (more imbalanced). In comparison, RankSVM has higher drop (for Macro-FM it is $18\%$ and for Micro-FM it is $17\%$) and multi-label NB has the highest drop (for Macro-FM it is $38\%$ and in Micro-FM it is $41\%$) for the same. These results indicate that overall LSTM-Att and RankSVM models are more robust against data imbalance.

 \begin{equation}
 \begin{split}
 \label{eq:7}
  PD(model) = (\frac{1}{AVG_{9}(model.BOW.FM, model.WE.FM)}\\ \times \{(AVG_{9}(model.BOW.FM, model.WE.FM) \\- (AVG_{16}(model.BOW.FM, model.WE.FM) \}) \times 100
  \end{split}
 \end{equation}
 
 \begin{equation}
 \begin{split}
 \label{eq:8}
 RankSVM-PD = AVG(PD(RankSVM-PPT), \\ PD(RankSVM-LP)) 
  \end{split}
 \end{equation}

\subsection{Confusion Matrices}
\begin{figure}[!ht]
    \centering
    \includegraphics[width=0.70\textwidth]{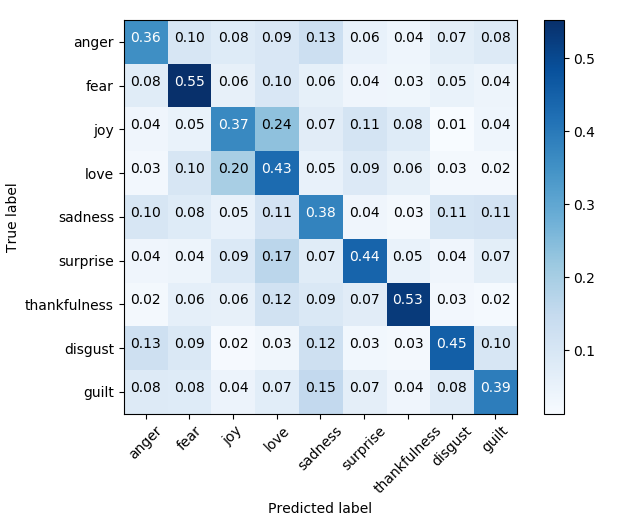}
    \caption{\label{fig:9conf} Confusion matrix for 9 emotions}
\end{figure} 

\begin{figure}[!ht]
    \centering
    \includegraphics[width=0.70\textwidth]{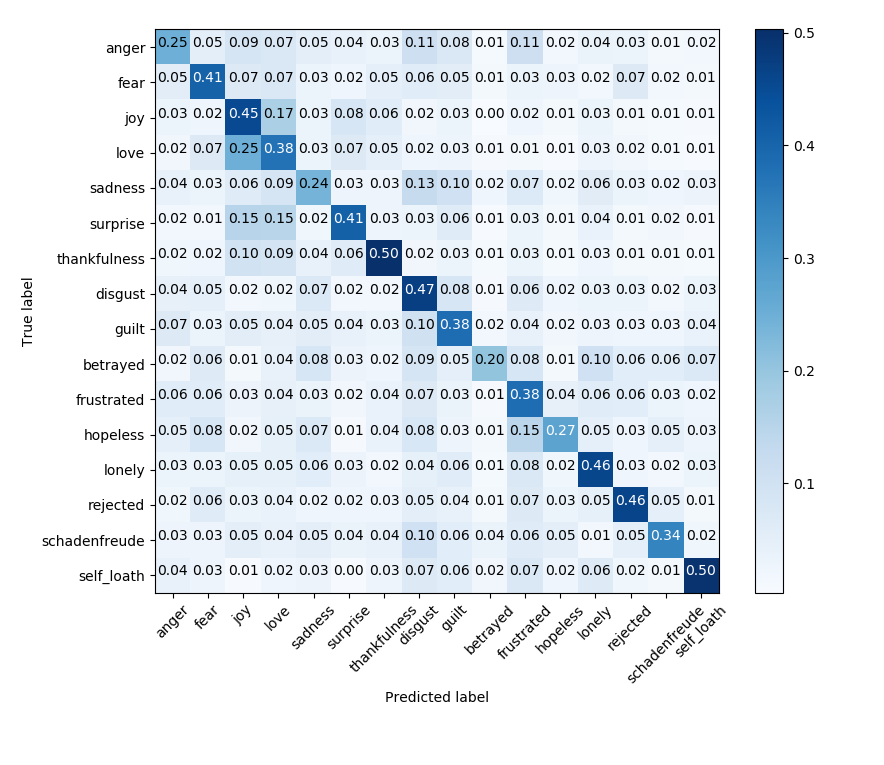}
    \caption{\label{fig:16conf} Confusion matrix for 16 emotions}
\end{figure} 


An ideal confusion matrix should be strictly diagonal with all other values set as zero. In our multi-label case, we see that our confusion matrix has the highest values along diagonal implying it is correctly classifying most of the emotions (Figure \ref{fig:9conf}. Non-diagonal values imply incorrect classification, where $love$ and $joy$ and $frustrated$ and $hopeless$ are mostly confused labels because these emotion labels tends to occur together.


\section{Conclusion and Future Work}

We have experimented with two state of the art models for a multi-label emotion mining task. We have described a detailed outline of data collection and processing for our two multi-label datasets, one containing Tweets with nine basic emotions 
and another having those Tweets augmented with additional Tweets from seven new emotions (related to depression). We also use two widely used features for this task, including bag-of-words and word embedding. Moreover, we provide a detailed analysis of these algorithms performance based on Macro-FM and Micro-FM measures. Our experiments indicate that a deep learning model exhibits superior results compared to others; we speculate that it is because of improved capture of subtle differences in the language, but we lack an explanatory mechanism to confirm this. Moreover, deep learning and RankSVM models are both better in handling data imbalance. It is also to be noted that word embedding feature based deep learning model is better than bag-of-words feature based deep learning model, unlike Na{\"i}ve Bayes and RankSVM models. As expected, this confirms that Deep Learning models are good with dense word vectors rather than very sparse bag-of-words features.
In the future, we would like to do a finer grained analysis of Tweets from depressed people, based on these extended emotions, and identify the subtle language features from the attention layers outputs, which we believe will help us to detect early signs of depression, to monitor depressive condition, its progression and treatment outcome. 

\bibliographystyle{splncs04}
\bibliography{nldb2019}
\end{document}